# Automatic Recognition and Classification of Future Work Sentences from Academic Articles in a Specific Domain

Chengzhi Zhang [1,2,*], Yi Xiang [1], Wenke Hao [1], Zhicheng Li [1], Yuchen Qian [1], Yuzhuo Wang [1]

1. Department of Information Management, Nanjing University of Science and Technology, Nanjing, 210094, China
2. Key Laboratory of Rich-media Knowledge Organization and Service of Digital Publishing Content Institute of Scientific & Technical Information of China, Beijing,100038, China

**Abstract:** Future work sentences (FWS) are the particular sentences in academic papers that contain the author's description of their proposed follow-up research direction. This paper presents methods to automatically extract FWS from academic papers and classify them according to the different future directions embodied in the paper's content. FWS recognition methods will enable subsequent researchers to locate future work sentences more accurately and quickly and reduce the time and cost of acquiring the corpus. At the same time, changes in the content of future work will be illuminated, and a foundation will be laid for a more in-depth semantic analysis of future work sentences. The current work on automatic identification of future work sentences is relatively small, and the existing research cannot accurately identify FWS from academic papers, and thus cannot conduct data mining on a large scale. Furthermore, there are many aspects to the content of future work, and the subdivision of the content is conducive to the analysis of specific development directions. In this paper, Nature Language Processing (NLP) is used as a case study, and FWS are extracted from academic papers and classified into different types. We manually build an annotated corpus with six different types of FWS. Then, automatic recognition and classification of FWS are implemented using machine learning models, and the performance of these models is compared based on the evaluation metrics. The results show that the Bernoulli Bayesian model has the best performance in the automatic recognition task, with the Macro $F_1$ reaching 90.73%, and the SCIBERT model has the best performance in the automatic classification task, with the weighted average $F_1$ reaching 72.63%. Finally, we extract keywords from FWS and gain a deep understanding of the key content described in FWS, and we also demonstrate that content determination in FWS will be reflected in the subsequent research work by measuring the similarity between future work sentences and the abstracts.

## 1 Introduction

Reading academic articles is a traditional way to assemble research ideas. In some special domains, for example, natural language processing, we often need to examine the problems the author failed to solve, the experimental methods that need to be improved, and the task performances that can be improved in the original articles, and use them as the direction of our follow-up research. Experienced researchers garner new ideas from the paper's theoretical explanations, experimental details, and so on. Furthermore, some researchers obtain this kind of information more directly, namely, by reading the future work of the paper. The future work section is an important part of a scientific article. The authors discuss extending their current work, approaches, or evaluations in the future work section (Hu & Wan, 2015). Sometimes, FWS can tell us the research starting point of a given research topic. However, it is true that some future work is vaguely written, as indicated in the Teufel study. The authors do not necessarily actually do it on purpose (Teufel, 2017); however, there is also content that is clearly expressed and can be promoted, such as: *for future research, to reduce the computational overhead, we will work on methods for sample selection as follows: introduction of constraints on nondependency* (Kudo & Matsumoto, 2000), and from the content of this sentence we know that the author's experiment in the original text can be improved by introducing constraints on nondependencies. The study of FWS can also assist in researching hot topic predictions. By acquiring large-scale future work sentence data in a specific field and classifying them according to their descriptions, we can receive a preliminary understanding of the research directions that scholars focus on most over time. On this basis, we can extract the most mentioned terms in combination with named entity extractions and other techniques, which can help in perceiving the subsequent hot research directions in the field.

---

[*] Corresponding author, Email: zhangcz@njust.edu.cn

Currently, there is a lack of investigation on FWS. One reason for the current paucity of research on future work is that such content is not as easily extracted from papers as the abstracts, keywords, headings, chapter structures, etc. The future work chapter is outside the standard academic paper IMR&D (Sollaci & Pereira, 2004) structure, which means that content describing future work is mixed in the discussion and conclusion sections (Zhu et al., 2019). Most of these sections talk about something other than future work. How to identify this section more accurately is the basis for any subsequent large-scale analysis, however, no effective solution is provided in some of the existing works. Therefore, there is first a need to identify future work from academic papers. In addition, by reading the future work content of a large number of papers in the domain of natural language processing, we uncover many different elements in the future work descriptions, some emphasizing ways to improve the algorithms and some discussing applying the models to larger datasets. Therefore, based on identifying future work, we need to classify multiple types of future work sentences, which can help us better understand the research focus of future work in this domain.

We refer to the sentences describing future work in academic papers as FWS and use sentences as the smallest unit to describe future work. This paper aims to automatically recognize and classify future work sentences from academic articles in a specific domain, namely, natural language processing (NLP). Our experimental data come from three major conferences in the NLP domain, the Association for Computational Linguistics (*ACL*), the Conference on Empirical Methods in Natural Language Processing (*EMNLP*) and the North American chapter of the Association for Computational Linguistics (*NAACL*). First, the papers published in these three conferences from 2000 to 2020 are obtained, and some FWS in the papers were identified by manual labeling as a training set. Then, other sentences are recognized through the use of machine learning methods. With that, we construct an FWS classification system by grounded theory and combine it with a pretraining language model to classify FWS into different types. Finally, we analyze the content described in FWS based on keywords, and discuss whether FWS are valuable for the prediction of the follow-up research directions.

All the data and source code of this paper are freely available at the GitHub website: https://github.com/xiangyi-njust/FWS.

## 2 Literature Review

The related works includes are from two perspectives, corpus construction and content mining for future work and automatic classification of academic text.

### 2.1 Corpus construction and content mining for future work

Currently, research on future work sentences mainly includes extraction and classification, construction of a corpus, and discussion of the relationship between future work sentences and other parts of the paper. Hu & Wan (2015) extracted future work sentences from academic texts based on regular expressions, and then defined four types of future work sentences and manually labeled them. Finally, future work sentences in different fields were classified by comparing the combination of different features and traditional machine learning models. Li & Yan (2019) identified future work sentences by setting a series of rules and extracting keywords from these sentences. Then, they matched keywords in the future work parts with keywords in the title and abstract texts, thus obtaining the conceptual connection between papers in different fields and their future work parts. Zhu, Wang, and Shen (2019) used a BERT model to extract future work sentences. Four types of future work sentences are determined by the hierarchical clustering method. The following three works are most directly related to our research. Hao et al. (2020) constructed a high-quality corpus of future work sentences from ACL papers during 1990-2015, and then they expanded the corpus to the ACL, EMNLP and NAACL papers from 2000-2019, and trained the machine learning models but did not perform an in-depth analysis of the results (Hao et al. 2021). Qian et al. (2021) explored the connection between future work sentences and tasks in the NLP domain.

Table 1. Related works of future work sentences

| Time | Author | Corpus size | Corpus source | Domain | Classification method |
|---|---|---|---|---|---|
| 2015 | Hu and Wan | 3,708 papers | ACL Anthology | Nature Language Processing | Rule matching, SVM |
| 2019 | Li and Yan | 1,672 papers | Science Advances | Science, Engineering, Ecology, Agricultural Science, Medicine, et al. | Rule matching |
| 2019 | Zhu et al. | 1,579 papers | JASIST | Information Science | BERT, Hierarchical clustering |
| 2020 | Hao et al. | 4,024 papers | ACL Anthology | Nature Language Processing | Grounded theory |
| 2021 | Hao et al. | 9,508 papers | ACL Anthology | Nature Language Processing | Naive Bayesian, BERT |
| 2021 | Qian et al. | 13,035 papers | ACL Anthology | Nature Language Processing | Naive Bayesian, BERT |

The current research on FWS is summarized in Table 1, and it shows that the differences in the studies of FWS mainly lie in the original corpus sources, tagging classification framework and automatic classification methods. Our study is based on Hao's corpus (Hao et al. 2021), and we revised the samples with labeling errors, based on which we improved the performance of the multi-classifications by training the SciBERT model (Beltagy, Lo, & Cohan, 2019) and further expanded the corpus to ACL, EMNLP and NAACL papers from 2000-2020, along with a detailed content analysis of the corpus.

## 2.2 Automatic Classification of Academic Text

Automatic recognition and multi-type classification of FWS is essentially a fine-grained text classification task, so it is necessary to sort out the existing research on the sentence classifications of the academic text. Because of the small corpus size and the immature computer technology in early research, scholars used the rule-based method to classify texts and obtained better results (Hayes et al., 1990; Kappeler et al., 2008). In recent years, a research paradigm characterized as "data-driven" has emerged, and more researchers tend to use machine learning models for text classification (Badie, Asadi, & Mahmoudi, 2018; Kilicoglu et al., 2018). Traditional machine learning algorithms usually need to design suitable feature sets according to the given data, models and tasks but the deep learning algorithms can integrate feature engineering and predictive learning into one model to implement end-to-end training. The existing research on sentence classification tasks has achieved excellent results (Gonçalves, Cortez, & Moro, 2018; Jiang et al., 2019). Regardless of which model is used for text classification, the text must be represented first because the text is composed of unstructured strings and cannot be directly understood by computers. Word2Vec, proposed by Mikolov et al. (2013), is the most classic distributed representation model and has been continuously improved by researchers, including Glove (Pennington et al., 2014) and FastText (Joulin et al., 2017). However, the word vectors obtained by these types of representation models are characterized in a fixed way, which could not address the polysemy phenomenon until Peters et al. (2018) proposed a context-related text representation method called ELMo. Then, BERT (Devlin et al., 2018), GPT-3 (Brown et al., 2020), and other pretraining language models were proposed successively, and they performed well on several typical tasks in the NLP field.

The rule-based method uses a predefined rule set to classify each sentence, which requires deep domain knowledge. The training samples used in the machine learning method need to be labeled manually, and the cumbersome feature engineering makes it difficult to extend to new tasks and cannot make full use of a large amount of training data. Researchers have explored deep learning to address the limitations created by manual features. It is a mainstream trend to use deep learning to address text classification in the era of big data.

## 3 Methodology

### 3.1 Framework of this study

The goal of this paper is to recognize FWS in academic papers and classify them into different categories. We construct a type system of FWS based on grounded theory and then train machine learning models to automatically identify and classify FWS based on the manually labeled data.

Figure 1 shows the overall research process, which is divided into three main steps. The first step is the construction of a training corpus, including the extraction of the chapters related to future work, the construction of annotation specifications and the labeling of future work sentences. The second step is model selection and sentence classification. A variety of machine learning and deep learning models are trained on the training corpus, and then we choose the model with the highest $F_1$ to automatically recognize and classify FWS in unlabeled chapters. The third step is a result analysis, including content analysis, effectiveness analysis and a type distribution analysis of future work sentences.

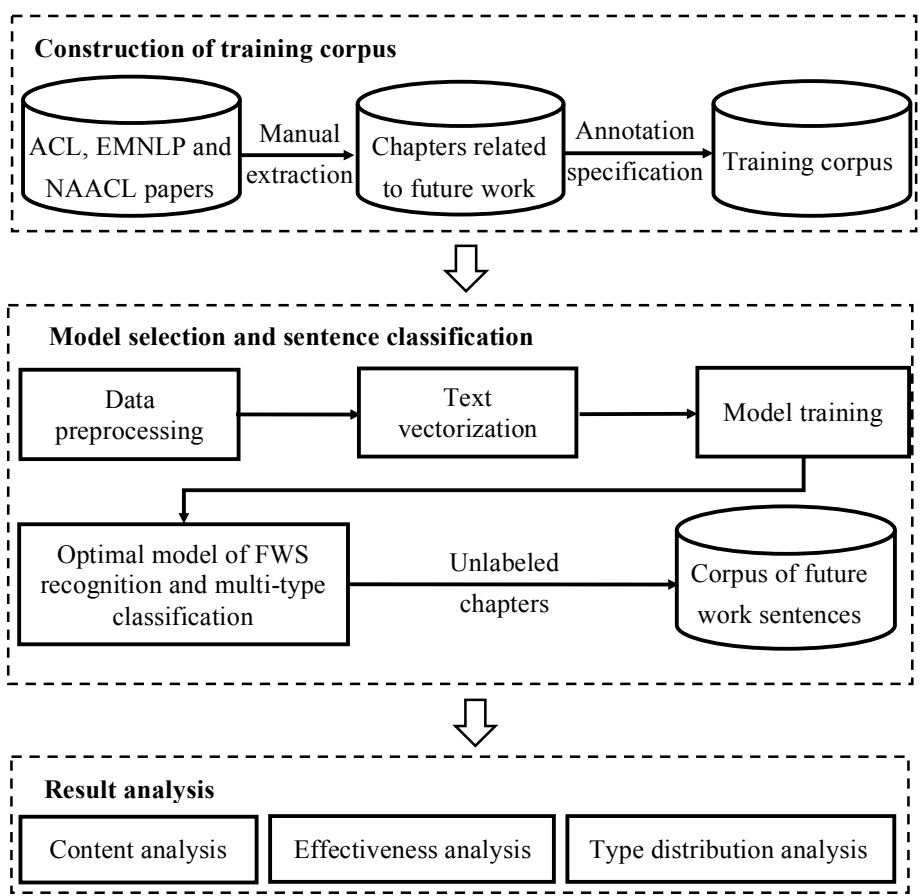

Figure 1. Framework of this study

## 3.2 Construction of annotation corpus

All the labeling tasks are completed by three senior undergraduates majoring in information management and information systems and reviewed by a professor of information science. ACL, EMNLP and NAACL are the top conferences in the field of NLP and are devoted to studying computing problems involving human language. From this, we collect 13,600 papers published by ACL, EMNLP and NAACL during 2000-2020 from the ACL anthology (https://www.aclweb.org/anthology/). First, the chapters related to future work are extracted manually to narrow the scope of subsequent annotations of FWS. Generally, authors explain their future work in "future work" chapters or "conclusion" chapters.

Therefore, if there is a "future work" chapter in a paper, we will directly extract the chapter. Otherwise, we will extract the "conclusion" chapter. Papers without these two chapters are removed from our corpus. The headings of these two chapters are written in various ways, and the headings are shown in Table 2 after unifying the case and lemmatization. Table 2 shows that future work sentences appear more in the Conclusion and Discussion chapters, and fewer articles use Future Work as a chapter alone.

Table 2. Frequency and ratio of the different chapter headings

| Headings | Frequency | Ratio |
| --- | --- | --- |
| Conclusion | 8376 | 70.31% |
| Conclusion and future work | 2274 | 19.09% |
| Discussion | 469 | 3.94% |
| Future work | 229 | 1.92% |
| Discussion and conclusion | 248 | 2.08% |
| Discussion and future work | 115 | 0.97% |
| Other[*] | 202 | 1.70% |

*Note*: Other refers to a chapter name that is not same as the other six types.

Then, future work sentence recognition and type labeling are conducted, and the labeling process is based on the ACL FWS-RC corpus (Hao et al., 2020). First, we analyze the corpus content, summarize the basic features of recognizing FWS, and annotate the beginning and end of FWS using <FW> and </FW> tags. Next, based on grounded theory, we classify FWS into the method, resources, evaluation, application, problem and other. We read the corpus word-by-word and sentence-by-sentence to determine the most important or most frequent initial concepts through sorting and analysis. Then, based on the potential logical relationship between the concepts, we gradually develop multiple subtypes and their main types. Finally, we conduct a saturation test to ensure that no new types are created. Table 3 shows the meaning of each FWS type and some examples (Hao et al., 2020).

Table 3. The meaning of each FWS type and some examples

| Type | Meaning | Examples |
| --- | --- | --- |
| Method | Optimization and improvement of models, features, algorithms and systems for the current study | *We plan to use argumentative zoning as a first step for IR and shallow document understanding tasks like summarization.* |
| Resources | Optimization, diversification, and expansion of resources used | *Future work involves scaling up to larger data and more features.* |
| Evaluation | Evaluate the results of current work or improve the means of evaluation | *We would like to explore other evaluation metrics (e.g., ROUGE-2, -SU4, Pyramid (Nenkova et al., 2007)).* |
| Application | Application of current research to other fields or tasks | *We are also planning to apply the proposed method to other tasks which need to construct tree structures.* |
| Problem | Research problems that have not been solved or that the author considers to be meaningful | *We suggest that sense matching may become an appealing problem and possible track in lexical semantic evaluations.* |
| Other | Future works that cannot be included in the above categories | *There are at least two potential future directions.* |

To ensure the accuracy of the labeling results during the labeling process, we randomly divide a subset from the original data; three annotators label this subset independently and then conduct a consistency test (Cohen, 1960). The results show that the Kappa values of FWS recognition are 0.752, 0.748 and 0.860, and the Kappa values of FWS classification are 0.709, 0.742 and 0.745, which shows that the labeling results are highly consistent; then the three annotators label the remaining data separately. Finally, sentences that cannot be evaluated are determined through an expert appraisal and group discussion.

The annotated corpus constructed in this paper consists of two parts. One part is the FWS labeled by ACL papers from 2000 to 2015 (Hao et al., 2020). Then, we add the annotation of FWS in the EMNLP and NAACL papers from 2000 to 2019. After manual labeling, we obtain a total of 9009 future work sentences. Frequency information of the corpus are shown in Table 4, from which it can be seen that the proportion of future work sentences in all the sentences is low. Combined with the information in Table 2, we can infer that this is because the future work sentences appear more in chapters such as Conclusion, while in the Conclusion chapters, the author is more concerned about summarizing and analyzing the article's experiments, and there is relatively little content that mentions future work. The other part is the multiclass labeled corpus. We label the 9009 future work sentences obtained in the previous stage, and the labeling results are shown in Figure 2. It can be seen that 4,912 sentences belong to the *method* type, accounting for more than half of the total FWS, while only 390 sentences belong to the *other* type. The difference in the number of sentences between the two types is approximately 14 fold, which shows that the corpus is unbalanced. The number of FWS of the *resources*, *application* and *problem* types is approximately 1000.

Table 4. Frequency information about the annotated corpus

| Conference | Period | Papers | Chapters | Future work sentences | Non-future work sentences |
|---|---|---|---|---|---|
| ACL | 2000-2015 | 3500 | 3,141 | 2,469 | 20,736 |
| EMNLP | 2000-2019 | 3893 | 3,608 | 4,161 | 21,450 |
| NAACL | 2000-2019 | 2639 | 2,264 | 2,379 | 13,701 |
| Total | 2000-2019 | 10032 | 9,013 | 9,009 | 55,887 |

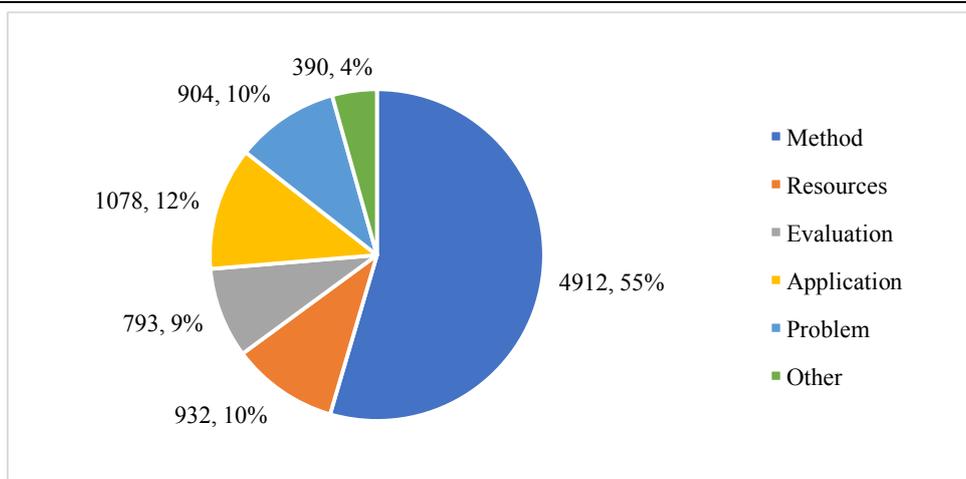

Figure 2. Number distributions of future work sentences in 6 different types

**3.3 Automatic recognition of future work sentences**

First, we preprocess the sentences in the annotated corpus, including removing symbols other than English letters and spaces, unifying word cases, and lemmatization. Then, we extract N-grams (N $\in$ [1,3]) from the sentences as features, representing the text as a vector in the feature space composed of these N-gram features, and the TF-IDF value (Salton et al., 1988) of the feature is used as the weight of the text vector. We then use traditional machine learning models for the recognition of future work sentences, including logistic regression (LR) (Genkin et al., 2007), naive Bayes (NB) (Kibriya et al., 2004), support vector machine (SVM) (Colas & Brazdil, 2006) and random forest (RF) (Breiman, 2001). Feature selection is also applied in our experiment because it usually leads to a better performance, lower computational cost and better model interpretability (Tang et at., 2014). In this experiment, we selected chi-square, a useful and reliable tool for the discretization and feature selection of numeric attributes (Liu & Setiono, 1995); we implement chi-square feature selection based on the scikit-learn package1. Finally, we can use the model with the highest $F_1$ value to recognize future work sentences in the unlabeled chapters.

**3.4 Automatic classification of future work sentences**

One of the divisions of future work sentences is the multi-classification problem since we have a total of 6 categories. Compared with the binary classification task, the semantic connection between each category in multi-classifications is more complicated; in addition, the sample size between each category in our labeled dataset is unbalanced. These problems indicate that it is difficult to use traditional machine learning models to obtain better classification effects. To solve the multi-classification problem, a deep learning model that can automatically extract text features and attain better text representation performance is our first consideration. However, when the amount of data is insufficient, it is not easy to show the advantages of the deep learning models. In our corpus, the sample size of each type is small, except for the method type, however, annotating more data manually is time-consuming and labor-intensive. So, we set our sights on the pretraining model; it has become a new research paradigm in the field of NLP to fine-tune downstream tasks based on pretraining models, which can help us achieve a better classification effect under limited data. BERT is a pretraining model developed by Google that relies on the encoder structure of Transformer and builds the basic model by stacking multilayer bidirectional transformers. Many past studies have demonstrated the effectiveness of BERT on multi-classification tasks (Mohammadi & Chapon, 2020).

    The original BERT was trained on a general corpus but our research object is scientific text, and it has common words that are quite different from the general corpus. Therefore, we use the SciBERT model, which follows the same architecture as BERT, except that it conducts the pretraining in scientific literature. Its corpus consists of 18% of papers in the field of computer science and 82% of papers in the field of biomedicine, and it uses the full text of the papers instead of just abstracts.

    For the multi-classification task of future work sentences, we directly use the SciBERT pretraining

---

[1] https://scikit-learn.org

model, and then input the future work sentences after preprocessing into SciBERT to fine-tune the model parameters. The SciBERT model consists of three layers. First, the sentence is converted into a vector in the input layer, and then the input vector is represented in the embedding layer to complete the mapping of the text sequence to a multidimensional vector space. Next, after the vectors pass through a dense layer, we obtain the final vector representation of each FWS. To obtain the classification results, we input the obtained feature vectors into the Softmax classifier (Mikolov et al., 2011) to calculate the probability distribution and determine the type with the highest corresponding probability to predict the type of future work sentences. From this, we obtain a language model relevant to our task. Finally, we input the unlabeled FWS into this model for prediction.

### 3.5 Content analysis of future work sentences

To understand the content described in FWS, we adopt a keyword extraction strategy and use keywords to describe the key research topics and directions in the future work. After comparing some mainstream methods, such as TextRank (Mihalcea et al., 2004) and Rake (Stuart et al., 2010), we found that the extraction effect is not excellent, and finally, we chose KeyBert[2]. KeyBert calculates the degree of correlation between candidate keywords and the document itself through BERT word embeddings and cosine similarities and then outputs the top-n keyword set with similarities. We use the KeyphraseVectorizer[3] method to extract candidate keywords. Unlike n-gram, which directly specifies the word range, KeyphraseVectorizer can extract phrases that conform to the grammatical rules from the text, avoiding the need to find the best n value in the n-gram method through comparative experiments but also ensuring a better extraction effect. In terms of the number of keywords in the final output, through manual reading, we find that there are fewer than five keywords with practical significance for a future work sentence. After further comparison experiments, we determined the n value of top-n to be 3. Then, we further process the extracted keyword set, including stemming and lemmatization, removing some high-frequency but meaningless words, such as "future work," and some words that are too broad, such as "machine learning" and "algorithm". Finally, the keywords of different years are sorted according to the word frequency, with the year as the dimension used for subsequent operations.

### 3.6 Effectiveness measure of future work sentences

The FWS in a scientific research paper describes the future prospects, such as the directions, hotspots, possible problems, and the corresponding solutions of the follow-up research. We extract FWS from papers and then extract the keywords. These processes allow us to grasp the key points of the future work described by the authors, however, they cannot explain the effectiveness of FWS, that is, whether they can be reasonably reflected in the subsequent research work.

To evaluate the effectiveness of FWS, this paper comprehensively considers the FWS in academic papers at a specific time, and the abstracts in other papers published several years later. The abstract is an overview of a paper's work, and the main content of the research can be surmised by extracting the keywords in the abstract. This paper assumes that if the future work sentences of the *kth* year can indeed provide researchers with new research directions, the keywords related to the description of the future work sentences of the *kth* year should appear in the abstract keyword set of the *(k+n)th* year. Based on this assumption, we obtain the paper's abstract, extract the keywords in the FWS and the abstract and compare the similarity between the two parts of the keyword set in units of years. The specific operations are as follows:

(1) Extract a set of keywords from the future work sentences of the kth year, denoted as $f_k$. $f_k = <(w_f^1, c_f^1), (w_f^2, c_f^2), \ldots, (w_f^i, c_f^i)>$, where $w_f^i$ represents the keyword, and $c_f^i$ represents the number of times the keyword appears in the kth year (more than one occurrence in a paper is only recorded once);

(2) Extract the keyword set from the abstracts of papers n (n>=1) years after the kth year, denoted as $a_{k+n}$. $a_{k+n} = <(w_a^1, c_a^1), (w_a^2, c_a^2), \ldots, (w_a^i, c_a^i)>$, where $w_a^1$ represents the keyword, and $c_a^i$ represents the number of times the keyword appears in the (k+n)th year.

(3) Take the union of the keywords of $f_k$ and $a_{k+n}$, denoted as W; $W = <w_f^1, w_f^2, \ldots, w_a^1, w_a^2, \ldots>$.

(4) Convert $f_k$ and $a_{k+n}$ into vectors $v_f^k$ and $v_a^{k+n}$, where each dimension of the vector represents a word of W, and the value in each dimension is the proportion of the occurrence of the word in the set to the total. Taking $f_k$ as an example, $w_f^1$ exists in $f_k$, and the proportion is $\frac{c_f^1}{\sum c_f^i}$, so the value of the corresponding dimension in $v_f^k$ is the proportion; $w_a^1$ does not exist in $f_k$, and the value of the corresponding dimension is 0;

(5) Calculate the cosine similarity between $v_f^k$ and $v_a^{k+n}$ to obtain the degree of correlation between the future work sentence of the kth year and the summary of the *(k+n)th* year. The calculation formula of the cosine similarity (Salton et al., 1983) is as follows:

$$cosine(f_k, a_{k+n}) = \frac{v_f^k \cdot v_a^{k+n}}{|v_f^k| * |v_a^{k+n}|} \quad (1)$$

---
[2] https://github.com/MaartenGr/KeyBERT

[3] https://github.com/TimSchopf/KeyphraseVectorizers

In terms of abstract data, we obtain the XML format data of the papers from ACL, EMNLP and NAACL, three natural language processing conferences from 2000 to 2015 from the ACL anthology reference corpus, then extract the abstracts from the papers and check them manually. For the abstract data from 2016 to 2021, we download the paper data in a PDF format from the ACL anthology website and extract the abstract part through a manual process. The keyword extraction method of the abstract is consistent with the processing method on FWS in Section 3.5.

# 4 Experiment and Results

## 4.1 Automatic recognition result of future work sentences

In the training of this experiment, we combine the learning curve and grid search to adjust the superparameters. There are two super parameters that need to be constantly adjusted to achieve the best model effect. One is used to set the number of features selected, and the other is used to set the super parameters inside the model. Another essential step in the training process of this experiment is to undersample the samples. Undersampling is a popular technique for unbalanced datasets to reduce the skew in the class distributions (Dal et al., 2015). The categories of positive (future work sentences) and negative samples in our original training set are unbalanced, and the number of negative samples is approximately six times that of the positive samples. In this case, the model's performance is affected mainly by negative examples, and the trained model may not have an exemplary discriminative performance for positive examples. Therefore, in the experiment, we undersampled the negative samples. We finally kept the number of negative samples consistent with the number of positive samples through experimental comparisons, there are 9009 samples for both categories. For the model's evaluation index, we use the macro average value of precision, recall and $F_1$. The evaluation results of each model are obtained based on a 10-fold cross-validation. The results are shown in Table 5.

Table 5. Automatic recognition performance of FWS (%)

| Model | Macro_Avg | | |
|---|---|---|---|
| | Precision | Recall | $F_1$ |
| LR | 86.42 | 85.89 | **86.15** |
| SVM | 87.68 | 87.41 | **87.54** |
| NB | 90.86 | 90.61 | **90.73** |
| RF | 84.95 | 84.42 | **84.68** |

It can be seen that the naive Bayes model achieves the best performance in automatic recognition of the future work sentences; the macro average reaches 90.73, which shows that our model can distinguish future work sentences from nonfuture work sentences very effectively. We further analyzed the reason behind this. When describing a future work sentence, the author often uses prompt words such as *future work* and *we plan*, which can be captured through the n-gram feature model. Therefore, a simple naive Bayesian model that uses n-gram as a feature can perform excellently on this task. Then, we take all the samples of the annotated corpus as training sets and input the sentences in the unlabeled chapters into the trained model. We extracted 4,651 future work sentences from the ACL from 2016 to 2020, and from the EMNLP papers from 2020. The related information is shown in Table 6. Cumulatively, there are 13,660 FWS from the three conferences from 2000 to 2020.

Table 6. Recognition results of FWS in unlabeled chapters

| Conference | Period | Papers | Chapters | FWS | Non-FWS |
|---|---|---|---|---|---|
| ACL | 2016-2020 | 2,786 | 2,149 | 3,501 | 10,198 |
| EMNLP | 2020 | 782 | 751 | 1,150 | 3,331 |
| NAACL | 2020[*] | 0 | 0 | 0 | 0 |
| Total | 2016-2020 | 3,568 | 2,900 | 4,651 | 13,529 |

*Note*: The NAACL meeting was not held in 2020, so the data for the year is 0.

### 4.2 Automatic classification result of future work sentences

In this experiment, the classical deep learning models (TextCNN and BiLSTM) and the original BERT model are selected as benchmark models to compare the superiority and robustness of the pretraining language model SciBERT used in this paper. First, we choose TextCNN (Kim, 2014), a classic model for text classification in deep learning, and BiLSTM (Schuster & Paliwal, 1997), a variant of recurrent neural networks. Word2Vec is used in these two model word embedding layers. In addition, we chose the English pretraining model provided by the original BERT. Because it can learn more semantic information between sentences with a large sample size in advance, it has significant advantages under many text multi-classification tasks (Zhou, 2020). We randomly divide the preprocessed annotated corpus into a training set, verification set and test set at a ratio of 8:1:1. For the model evaluation index, we use the macro and weighted average values of precision, recall and $F_1$. The experimental results are shown in Table 7.

Table 7. Automatic classification performance of FWS (%)

| Model | Macro_Avg | | | Weighted_Avg | | |
|---|---|---|---|---|---|---|
| | Precision | Recall | $F_1$ | Precision | Recall | $F_1$ |
| BiLSTM | 50.63 | 47.62 | **47.49** | 63.64 | 61.64 | **61.87** |
| TextCNN | 54.26 | 51.55 | **52.57** | 65.17 | 65.63 | **65.31** |
| BERT | 71.20 | 60.43 | **62.38** | 74.48 | 72.70 | **72.12** |
| SciBERT | 69.50 | 61.24 | **63.03** | 73.85 | 73.70 | **72.63** |

The macro average is used to average the performance of the indicators in each category directly, and the weighted average is used to perform a weighted average of the indicator scores according to the sample size of each category. The more samples there are, the higher the weight of the category. From Table 7, we can see that all models have higher index performances on the weighted average than on the macro average, which indicates that the classification effects on different categories are quite different. According to the weighted average $F_1$ value, the classification result is TextCNN > BiLSTM. However, their $F_1$ value does not exceed 70%, which may be because the static word vector embedding model cannot deeply learn and capture the semantic connotation information of the sentences. BERT performs well overall, and this result may be related to the theoretical advantages of the BERT algorithm. BERT captures the global information of the text with the understanding of the context, so it can perform well in sentence-level tasks. Compared with BERT, SciBERT has a slight improvement in classification performance since the SciBERT model has the pretraining basis of the domain characteristics as a guarantee, which makes this model capture the most appropriate semantic information from each sentence.

Table 8. Automatic classification performance of SciBERT in each type

| Type | Precision | Recall | $F_1$ |
|---|---|---|---|
| Method | 79.18% | 86.23% | 82.56% |
| Resources | 54.87% | 66.67% | 60.19% |
| Evaluation | 66.04% | 85.37% | 74.47% |
| Application | 77.78% | 56.25% | 65.28% |
| Problem | 63.16% | 33.33% | 43.64% |
| Other | 76.00% | 39.58% | 52.05% |
| Macro_avg | 69.50% | 61.24% | 63.03% |
| **Weighted_avg** | **73.85%** | **73.70%** | **72.63%** |

Table 8 shows the classification effect of SciBERT on different categories. The *method* class has the best recognition effect due to having the most significant number of training samples, and the $F_1$ can reach 82.56. However, the classification performance is not always better with more training samples. A counterexample is the *problem* class, and the sample size of the *problem* training set is more than twice that of the *other*, but its $F_1$ value is lower than that of the *other* class, and is even less than 50%. The same is true for the *resources* class, and its accuracy is much lower than that of the other categories.

Then, we use the trained SciBERT model to automatically classify the unlabeled future work sentences extracted from Section 4.1. To improve the accuracy of subsequent analyses, the annotators manually check and correct the automatic classification results. Finally, we obtain a complete future work sentence corpus from 2000 to 2020. The number of different categories of FWS is shown in Figure 3. The number of future work sentences in the *method* type of the three conferences is the largest, while the number of future work sentences in the *other* type is the smallest. In the remaining four types, there is no obvious difference among the three conferences except that EMNLP has a slightly higher number of future work sentences in the *application* and *problem* type.

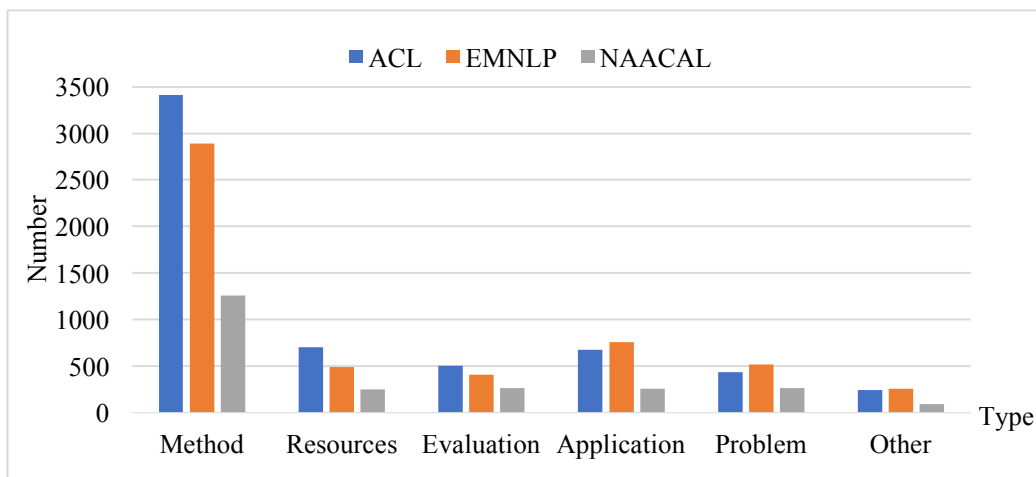

Figure 3. Number of different categories of FWS in the three conferences

## 4.3 Content analysis result of future work sentence

In this section, we count the high-frequency keywords to understand the content focus of the different FWS, such as which methods may be used in the future. To further analyze the behavior orientation of each keyword, the verbs adjacent to keywords are also extracted. Due to the limited space, the number of FWS of the method type accounts for more than half of all the sentences, which shows that researchers in the field of NLP pay more attention to methods, so the following takes the method type as an example.

Figure 4 lists the top 10 verbs with the highest frequency, among which "use" appears most frequently, and the FWS of the method type containing this verb includes "We encourage future work to judge model interpretability using the proposed evaluation and publicly published annotations", "Possible future directions include using more sophisticated feature design and combinations of candidate retrieval methods", "Future work includes using these approaches to induce model structure." and so on. The keywords near the verb may be the objects the authors will use or explore in the future.

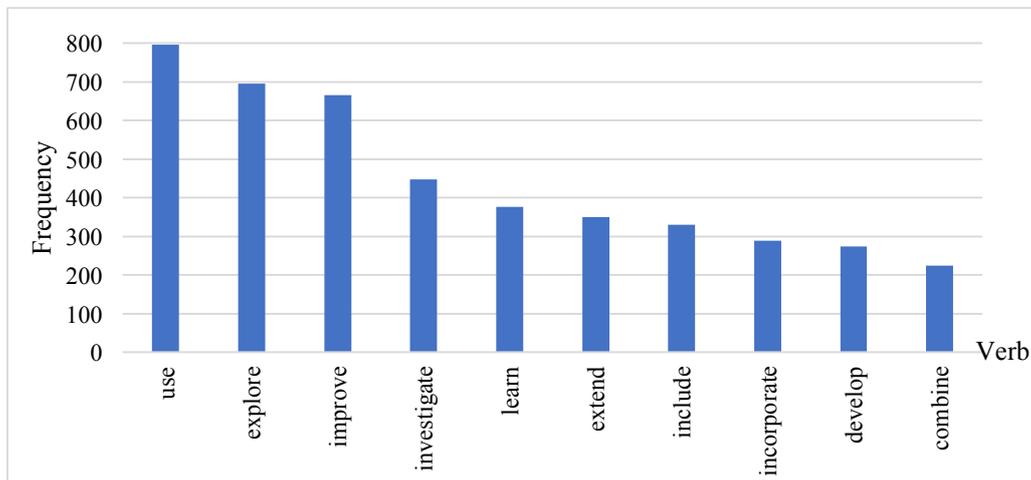

Figure 4. High-frequency adjacent verbs of keywords in FWS of *method* type

Table 9 shows the corresponding frequency of the top 10 high-frequency keywords and their adjacent verbs in the FWS of method type. The values in brackets represent the ratio of the frequency of occurrence of this verb to all the adjacent verbs of this keyword. From the high-frequency keywords, translation tasks and related models are mentioned most frequently. The rapid development of machine learning and deep learning has drawn new solutions to machine translation, and researchers use them to improve the machine translation quality. A large number of mentions of the embedding and language models benefit from the paradigm shift of natural language processing brought by BERT; the pretrained language model is undoubtedly one of the most popular research directions in the NLP field, constantly refreshing the SOTA level of many tasks in the NLP field. In addition, from these words, we can also discover popular current research methods in the NLP field, such as neural networks and graphs; research objects, such as entities; and research tasks, such as classification and generation. From the adjacent verbs, researchers often hope to use (use, train) or combine these methods to improve (improve, extend, develop) a task's performance or to improve these methods themselves.

Table 9. High-frequency keywords in FWS of the *method* type

| Keywords | Frequency | Adjacent verbs and their proportions |
|---|---|---|
| translation | 203 | use(4.9%); improve(4.6%); explore(3.2%); incorporate(2.5%); extend(2.3%); learn(1.5%); base(1.5%); investigate(1.4%); combine(1.3%); integrate(1.3%) |
| embedding | 190 | use(5.3%); embedded(4.8%); explore(4.8%); learn(3.1%); extend(1.9%); include(1.7%); combine(1.6%); investigate(1.6%); improve(1.6%); base (1.5%) |
| classification | 188 | use(5.2%); improve(3.3%); explore (2.6%); include(2.1%); train(2.1%); combine(1.8%); learn(1.8%); extend(1.7%); investigate(1.4%); integrate (1.2%) |
| graph | 178 | use(4.2%); explore(3.5%); improve(3.5%); extend(2.4%); investigate(2.3%); learn(2.1%); incorporate(2.0%); model(1.8%); consider(1.7%); include(1.6%) |
| language model | 175 | use(5.7%); improve(3.8%); explore(3.0%); investigate(2.0%); model(1.9%); learn(1.6%); extend(1.5%); generate(1.4%); develop (1.3%); combine(1.3%) |
| parser | 151 | use(5.3%); explore(3.6%); improve(3.2%); parse(2.4%); develop(2.3%); extend(2.1%); include(2.1%); base(1.5%); learn(1.5%); investigate(1.5%) |
| generation | 142 | explore(3.6%); use(3.5%); improve(3.3%); incorporate(2.8%); generate(2.7%); extend(1.8%); include(1.8%); learn(1.6%); train(1.5%); integrate (1.4%) |
| entity | 129 | improve(3.9%); use(3.0%); explore(2.1%); model(2.1%); investigate(1.9%); extend(1.8%);learn(1.8%);extract(1.5%); link(1.5%); develop(1.5%) |
| neural net | 100 | explore(3.8%); use(3.6%); improve(2.2%); investigate(2.0%); learn(1.8%); model(1.8%); develop(1.8%); base(1.4%); consider(1.4%); train(1.4%) |
| encode | 97 | encode(4.8%); use(4.0%); improve(3.3%); explore(2.9%); include(2.4%);incorporate(1.4%);propose(1.2%);introduce(1.2%); learn(1.2%); intend(1.2%) |

**4.4 Effectiveness result measures of future work sentences**

In this section, we first show the results of the similarity measure between FWS and the abstracts. Then, we illustrate the complex relationship between what is described in FWS and the actual situation with an example of a specific year. Finally, we elaborate on the shortcomings of the current use of FWS to make research hotspot predictions.

**(1) Results of similarity measures for FWS and Abstract**

Figure 5 shows the degree of correlation between the FWS for a given year and the abstract n years later. The horizontal axis represents the period, i.e., n years later; the vertical axis represents the corresponding year of the FWS; the deeper the cell in the heatmap, the higher the correlation. Longitudinally, the correlation between the abstracts of FWS and subsequent years becomes higher as time goes on. The maximum similarity is only approximately 0.24 by 2000, and reaches 0.86 by 2020, which indicates the relationship between FWS and the abstracts. The gradually increasing similarity reveals the practical value of FWS, and we can predict the focus and direction of subsequent studies based on the keywords of FWS.

Horizontally, we discover an interesting phenomenon. In some years, the correlation between FWS and abstracts decreases as the value of n increases; for example, the similarity between the FWS in 2000 and the abstract one year later is 0.19, and after 21 years, the similarity with the abstracts in 2021 is only 0.03; however, there are also some years in which the similarity decreases first as the value of n increases, however, it starts to increase again after reaching a minimum value, for example, in 2013. We analyzed the reasons behind these two cases. The former is easy to understand, because unless a field has been stagnant, it is almost impossible to predict the focus of the work in a decade or so based on the research directions proposed in one year of future work, especially for NLP, a field that is growing extremely fast, which is why the former case occurs mainly in earlier years; the latter is a more interesting phenomenon, which we analyzed in our case study with 2013 as an example.

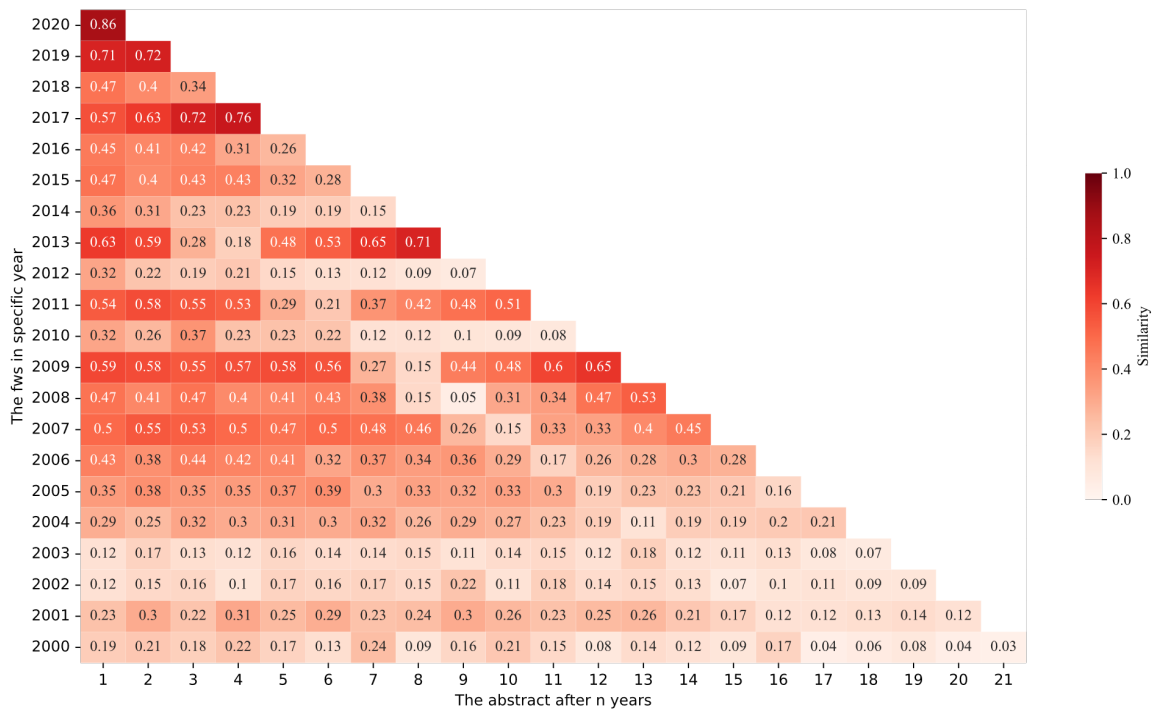

Figure 5. The degree of similarity between FWS and the abstract

**(2) Case study**

We use 2013 as an example to explain that the similarity between future work sentences and abstracts in some years in Figure 5 first decreases with increasing n values and then suddenly rises to a very high value. Table 10 gives the top 10 keywords with the highest frequency on the future work sentences in 2013, while the top 10 keywords of the abstracts in the representative years 2014, 2017 (the smallest point of similarity), and 2021 (the largest point of similarity) are selected according to the similarity values in Figure 5. First, it can be seen that no identical words appear in the FWS of 2013 and the abstracts of 2017, so the similarity between the vectors constructed based on the keywords would be low. Second, there are three repeated words in 2014 and only one in 2021, but the similarity between the abstracts of 2021 and the future work sentence is 0.71, while it is only 0.63 in 2014. Because we use the weight information when calculating the similarity, the greater the proportion of the number of times a keyword appears in the year, the greater the weight will be. In 2021, the frequency of the *language model* is much higher than that of the other words, so the corresponding weight in the keyword vector of that year is greater, while in 2014, although there are several words repeated, the weight of these words in the overall frequency is not very high, so in the calculation of similarity, 2021 will actually be greater.

We analyzed the reason behind this phenomenon. Mikolov et al. proposed the Word2vec model in 2013 (Mikolov et al., 2013). Word2vec learns the relationship between words by training neural networks and maps the words into vectors to obtain a text representation, which is a better way to model language compared to the Bag-of-Words or N-grams models. The proposal of Word2vec caused the language model to receive increased attention in 2013 and 2014, however, in the following years, the language model did not make further advancements, while the convolutional neural networks, recurrent neural networks, LSTM, and other network structures made rapid advancements in these years, especially with the emergence of the Transformer architecture in 2017 (Vaswani et al., 2017), making the attention mechanism in the field of NLP popular. So, in the top 10 keywords in 2017, we see more mainstream network structures at that time. A change occurred in 2018, when the Google team proposed BERT, which refreshed SOTA on many NLP tasks, and pretraining + fine-tuning also became a new research paradigm in the NLP field; from 2018 to now, pretrained language models have been the hottest research topic in the NLP field, so the language model in the top 10 keywords in 21 years is once again in the first place.

In summary, for cases where the similarity between future work sentences and the abstracts decreases and then increases, as represented by 2013, we surmise that it is because a research topic mentioned in the past has attracted widespread attention in the field again by new research ideas after many years. The analysis of such cases helps us to have a clearer understanding of the development paths of the algorithms and models in the field.

Table 10. High-frequency keywords in the FWS of 2013 and the Abstract of 2014, 2017 and 2021

| FWS(2013) \ Abstract | 2014 | 2017 | 2021 |
|---|---|---|---|
| **language model** | **language model** | neural machine translation | **language model** |
| **classification** | static machine translation | word embedding | bert |
| entity | translate quality | neural network | neural machine translation |
| parser | knowledge base | recurrent neural network | attention |
| **machine translation** | topic model | decode model | knowledge graph |
| speech tag | **machine translation** | convolutional neural network | entity recognition |
| summary length | entity | attention mechanism | knowledge base |
| distribution semantics | Tweet | encode | neural network |
| language pair | **classification** | lstm | transform |

| word alignment | wordnet | sequence model | embedding |

*Note: The bolded words in the column refer to the keywords that overlap in the future work sentences and abstracts.*

The conclusions of this section are not sufficient to prove a causal relationship between the FWS and the follow-up research content. We cannot support the conclusion that if some specific directions are mentioned in future work, then the follow-up research content will definitely develop in these directions. The most critical point of this section's research is to prove a relationship between the two parts. To determine whether it is a causal or another type of relationship, a follow-up analysis of the data is needed.

## 5 Discussion

Our work contributes to constructing a large-scale corpus using a combination of quantitative and qualitative methods, with a reliable corpus to discuss future research in the NLP field. The above results clearly describe the basic composition and relevant quantitative information of the corpus. In this section, we first discuss the changes in FWS types combining the existing results with time, and then summarize the important findings of our study.

**5.1 Future research trends reflected by the evolution of the FWS type**

To observe the overall trend of future research in the NLP field, we calculate the percentage of the number of FWS in each type per year and plot the evolution diagram, as shown in Figure 6. Each year's portion of the method type has apparent advantages, and the remaining five classes with the year exhibit small-scale volatility. From 2015 to 2020, the percentage of the *method* type decreased while the percentage of the *application* type increased but it was still small compared with *the method* type. Based on the results, we can infer that the overall trend of future research on NLP should still focus on the methods.

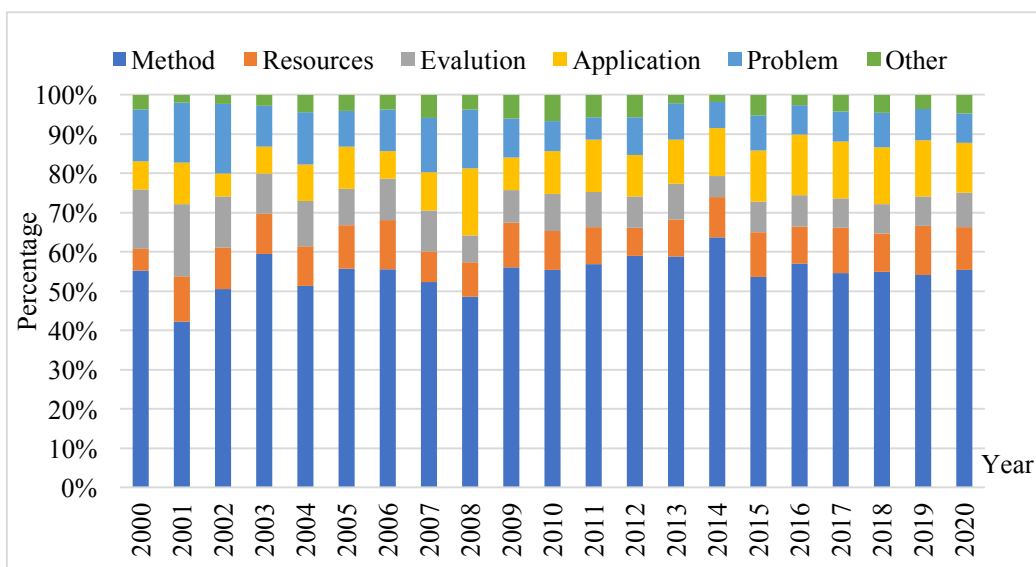

Figure 6. Percentage of FWS types from 2000 to 2020

ACL, EMNLP and NAACL are all top conferences in the NLP field but their emphasis varies. For example, ACL is mainly oriented to research related to computational linguistics theory, while EMNLP is more about the academic exploration of natural language algorithm solutions in different fields. Therefore, it is imperative to conduct an evolution of FWS types for the three conferences. It is worth mentioning that the annual number of FWS is easily affected by the expansion or contraction of the total number of papers. To correct for random fluctuations, we use the ratio of the number of papers containing FWS of a certain type to the total number of papers as an indicator. As seen from Figures 7-9, the ratio of the *method* type is the largest in most years for all three conferences, indicating that researchers are more inclined to propose methodological improvements in future work. Some differences are found by calculating the average values. The ratio of the method type is 0.29 for ACL, 0.26 for EMNLP and 0.31 for NAACL, which indicates that NAACL places more emphasis on the discussion of methods than the other two conferences. The other five types of FWS are distributed with no significant differences.

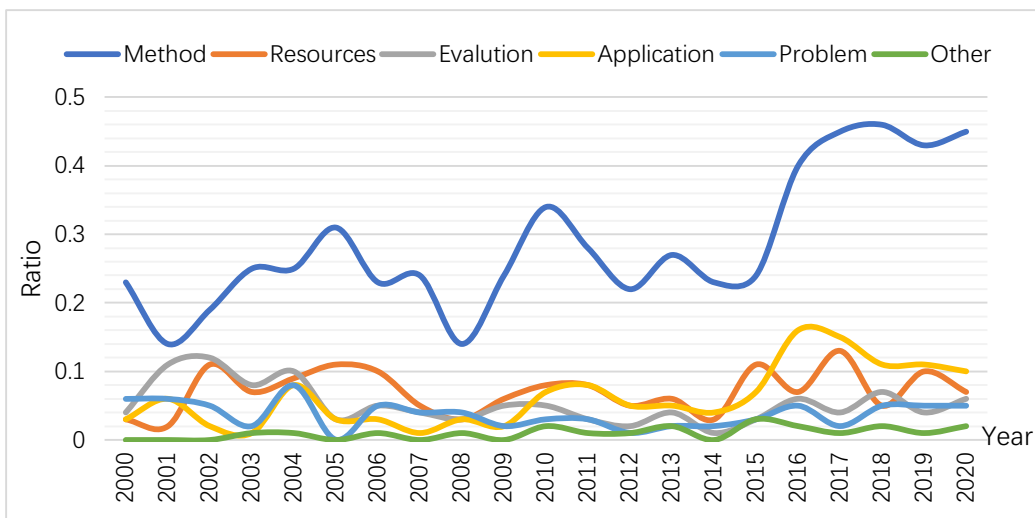

Figure 7. Evolution of FWS types in ACL

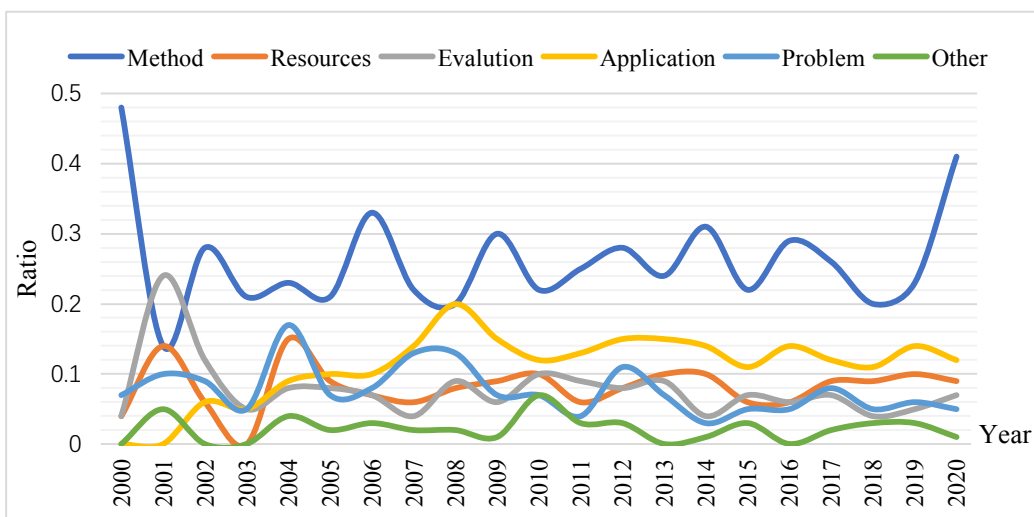

Figure 8. Evolution of FWS types in EMNLP

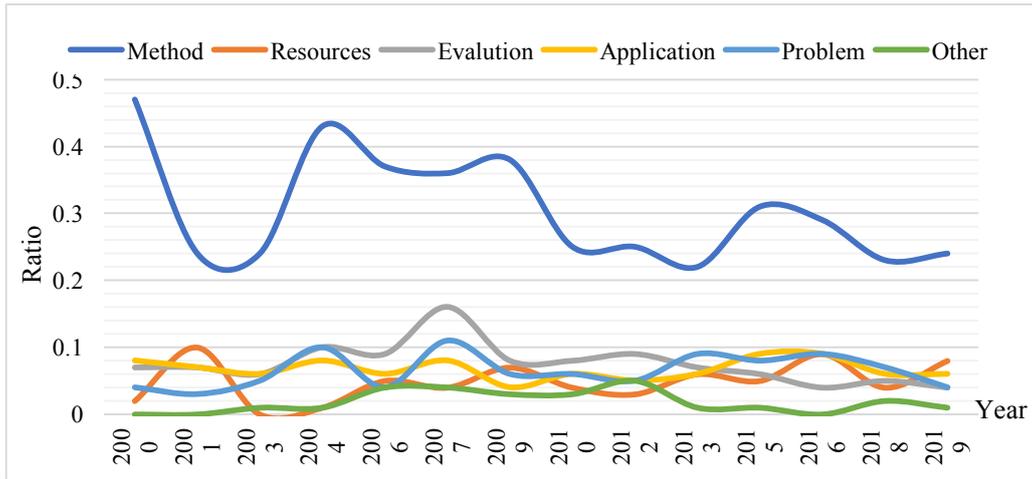

Figure 9. Evolution of FWS types in NAACL

After discussing the future research trend of the macro field, we can focus on the future trend of specific tasks. Due to space limitations, we only show and analyze the results for the task of machine translation since it is the one that appears most frequently. As shown in Figure 10, machine translation is most often mentioned in the FWS of the *method* and *application* types. Overall, the percentage of FWS in these two types fluctuates up and down with the years, with no significant and consistent upward or downward trend. However, the percentage of the *application* type has increased in recent years, surpassing the percentage of the *method* type after 2015. We suspect that with the maturity of machine translation technology, it is gradually being applied to various NLP tasks.

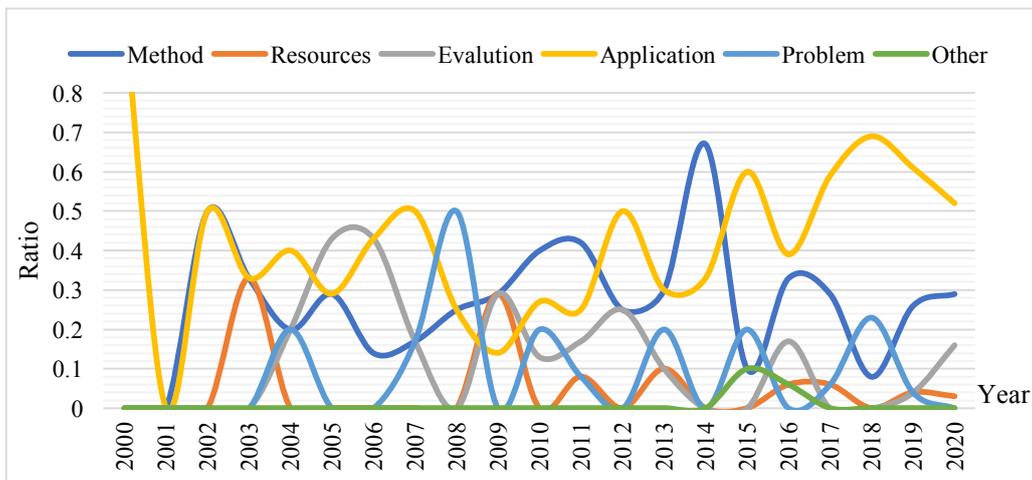

Figure 10. Evolution of FWS types in machine translation

### 5.2 Important findings

Some important findings are obtained through our research:

First, our experiment's results show that the four traditional machine learning models are similar in their performance of the automatic recognition task, and applying the feature selection method can significantly improve the model's effect. For the automatic multi-classification task, the SciBERT model outperforms the other baseline models in most metrics, although training takes a longer time. We solve the problems existing in manual annotation, including that rule-based methods have difficulty covering all the language features in the future work sentences and are susceptible to subjective factors. More sophisticated models should be explored to improve the performance in the future.

Second, the content analysis on future work sentences shows that machine translation tasks are the most studied in the NLP field; in addition, language models are also a hot topic of research in the NLP field in recent years; from the verbs co-occurring with keywords, we find that the current NLP field focuses more on how to use and improve the algorithms or models. In the analysis of the effectiveness of future work sentences, we found that the similarity between future work sentences and the abstracts has become very high in recent years from the perspective of keywords, which means that it is feasible to use future work sentences for the prediction of subsequent research focuses, and we gave a preliminary explanation the algorithm and model development trends in the field reflected by the change in similarity through a case study.

## 6 Conclusions and Future Work

In this paper, we focus on the recognition and multi-classification of future work sentences, and mine the research focus reflected in the future work sentences. A training set of future work sentences is constructed by manual tagging. Then, automatic recognition and multi-classification of future work sentences are achieved by using machine learning and deep learning models, and an unlabeled corpus is input into the optimal model, thus obtaining a corpus of future work sentences of the ACL, EMNLP and NAACL conferences during 2000-2020. On this basis, we extracted keywords in future work sentences. By counting the frequency of the keywords, we further understand the key tasks and important research directions in the NLP field for the future. By calculating the similarity between the keywords of the future work sentence and the abstract n years later, our preliminary exploration of the link between future working sentences and subsequent research priorities can provide an experimental basis for more in-depth analysis of how future work sentences influence subsequent scientific research, and how they can be used for tasks such as technology prediction.

There are still some limitations in this study. First, the dataset we used for model training is domain specific, and it is unclear whether it is suitable for the extraction and multi-classification of FWS in other fields. Second, we established six categories for FWS, *method*, *resources*, *evaluation*, *application*, *problem* and *other*, but the granularity of the classification is not detailed enough. For example, methods can be further divided into the grounded theory method, bibliometrics method, machine learning method, etc. Third, this paper only uses keywords to measure the degree of correlation between the future work sentences and the abstracts, and does not calculate the similarity between sentence levels and deeper semantics. Based on these findings, our future work is as follows.

At present, the supervised deep learning model is the mainstream method for automatic text classification, however, there is no guarantee that the FWS classification model used in this paper will apply equally well to data from other domains, and it is time-consuming to label new data. Therefore, it is a future research direction to use semisupervised or unsupervised methods to reduce the manual labeling and achieve similar results. In addition, a practical application of FWS is to predict research trends. However, a simple word frequency analysis is not sufficient to accurately predict future research trends, so we plan to conduct a deeper semantic analysis based on the context, including classifying the task type of the article where the future work sentence is located. By determining the task type of the article, we can gain a clearer grasp of the directionality of the content reflected in the future work sentences.

## Acknowledgment

This study is supported by the National Natural Science Foundation of China (Grant No. 72074113) and the opening fund of the Key Laboratory of Rich-media Knowledge Organization and Service of Digital Publishing Content (Grant No. zd2022-10/02).